\documentclass[singlespacing, twocolumn]{bmcart}

\usepackage{etoolbox}
\makeatletter
\patchcmd\set@numberlines@box{\rlap}{\@gobble}{}{}
\makeatother

\usepackage{float, amsthm, amssymb, amsmath, bm, graphicx, multirow, url, stfloats, etoolbox, boldline, rotating, array, cleveref, mathtools, multirow, ragged2e}

\theoremstyle{plain}

\newtheorem{definition}{Definition}

\RequirePackage[numbers]{natbib}
\usepackage[utf8]{inputenc} 

\startlocaldefs
\endlocaldefs

\begin{document}
	
\begin{frontmatter}

\begin{fmbox}
\dochead{Research}


\title{Sort \& Slice: A Simple and Superior Alternative to Hash-Based Folding for Extended-Connectivity Fingerprints}


\author[
	addressref={aff1},                                    		
	email={markus.dablander@maths.ox.ac.uk}
]{\inits{M.D.}\fnm{Markus} \snm{Dablander}}
\author[
	addressref={aff3},
	email={Thierry.Hanser@lhasalimited.org}
]{\inits{T.H.}\fnm{Thierry} \snm{Hanser}}
\author[
	addressref={aff1},
	email={renaud.lambiotte@maths.ox.ac.uk}
]{\inits{R.L.}\fnm{Renaud} \snm{Lambiotte}}
\author[
	addressref={aff2},
	corref={aff2},
	email={garrett.morris@stats.ox.ac.uk}
]{\inits{G.M.M.}\fnm{Garrett M.} \snm{Morris}}


\address[id=aff1]{%
	\orgdiv{Mathematical Institute},
	\orgname{University of Oxford},
	\street{Andrew Wiles Building, Radcliffe Observatory Quarter (550), Woodstock Road},
	\postcode{OX2 6GG,}
	\city{Oxford},
	\cny{United Kingdom}
}

\address[id=aff3]{%
	\orgname{Lhasa Limited},
	\street{Granary Wharf House, 2 Canal Wharf},
	\postcode{LS11 5PS,}
	\city{Leeds},
	\cny{United Kingdom}
}

\address[id=aff2]{%
	\orgdiv{Department of Statistics},
	\orgname{University of Oxford},
	\street{24-29 St Giles’},
	\postcode{OX1 3LB,}
	\city{Oxford},
	\cny{United Kingdom}
}





\begin{abstractbox} 

\begin{abstract} 
\justifying

Extended-connectivity fingerprints~(ECFPs) are an indispensable and ubiquitous tool in current cheminformatics and molecular machine learning. Alongside graph neural networks and physicochemical descriptors, ECFPs are one of the most prevalent molecular feature extraction techniques used for chemical prediction. Atom features learned by graph neural networks can be aggregated to compound-level vector representations using a large spectrum of available graph pooling methods; in contrast, sets of detected ECFP substructures are by default transformed into bit vectors using only a simple canonical hash-based folding procedure. We introduce a general mathematical framework for the vectorisation of structural fingerprints by defining a formal operation called \textit{substructure pooling} that encompasses hash-based folding, algorithmic substructure-selection, and a wide variety of other potential techniques. We go on to describe \textit{Sort \& Slice}, an easy-to-implement and bit-collision-free alternative to hash-based folding for the pooling of ECFP substructures. Sort \& Slice first sorts ECFP substructures according to their relative prevalence in a given set of training compounds and then slices away all but the $L$ most frequent substructures which are subsequently used to generate a binary fingerprint of desired length, $L$. We conduct a series of rigorous computational experiments to compare the predictive performance of hash-based folding, Sort \& Slice, and two advanced supervised substructure-selection schemes (filtering and mutual-information maximisation) for ECFP-based molecular property prediction. Our results indicate that, despite its technical simplicity, Sort \& Slice robustly (and at times substantially) outperforms traditional hash-based folding as well as the other investigated substructure-pooling methods across distinct molecular property prediction tasks, data splitting techniques, machine-learning models and ECFP hyperparameters. We thus recommend that Sort \& Slice canonically replace hash-based folding as the default substructure-pooling technique to vectorise ECFPs for supervised molecular machine learning. \textbf{Scientific contribution:} A general mathematical framework for the vectorisation of structural fingerprints called \textit{substructure pooling}; and the technical description and computational evaluation of \textit{Sort \& Slice}, a conceptually simple and bit-collision-free method for the pooling of ECFP substructures that robustly and markedly outperforms classical hash-based folding at molecular property prediction.

\end{abstract}


\begin{keyword}
\kwd{Extended-connectivity fingerprints}
\kwd{Morgan fingerprints}
\kwd{Molecular property prediction}
\kwd{Supervised machine learning}
\kwd{Hashing}
\kwd{Hash-based folding}
\kwd{Sort \& Slice}
\kwd{Feature extraction}
\kwd{Feature selection}
\kwd{Substructure pooling}
\end{keyword}


\end{abstractbox}
\end{fmbox}

\end{frontmatter}




\section*{Introduction}

\subsection*{Extended-Connectivity Fingerprints and Hash-Based Folding}

The use of extended-connectivity fingerprints~(ECFPs) is widespread in current cheminformatics and molecular machine learning. A modern and widely recognised technical description of ECFPs was given in $2010$ by Rogers and Hahn~\cite{rogers2010extended}, although the key ideas underlying ECFP-generation were introduced by Morgan~\cite{morgan1965generation} in $1965$. ECFPs have for instance been used successfully for ligand-based virtual screening~\cite{riniker2013open}, the prediction of the aqueous solubility of molecular compounds~\cite{duvenaud2015convolutional}, the computational detection of cytotoxic substructures of molecules~\cite{webel2020revealing}, the identification of binding targets of chemical compounds via similarity searching~\cite{alvarsson2014ligand}, and the prediction of quantum-chemical properties of small molecules~\cite{gilmer2017neural}. One of the most typical use cases of ECFPs is as a feature-extraction technique for supervised molecular machine learning, i.e.,~as a method to convert molecules into binary feature vectors for a given downstream molecular property prediction task. For this purpose, ECFPs are popular featurisations thanks to their conceptual simplicity, high interpretability, and low computational cost. Moreover, a considerable corpus of recent literature suggests that ECFPs still regularly match or even surpass the predictive performance of differentiable feature-extraction methods based on state-of-the-art message-passing graph neural networks~(GNNs)~\cite{stepivsnik2021comprehensive,mayr2018large,menke2021using,chithrananda2020chemberta,winter2019learning,dablander2023exploring,dablander2021siamese}.

At its core, the ECFP algorithm can be formally described as a method, $\varphi$, that transforms an input molecule, $\mathcal{M}$, often represented as a SMILES string~\cite{weininger1988smiles}, into a set of hashed integer identifiers
$$\varphi(\mathcal{M}) = \{\mathcal{J}_{1}, ..., \mathcal{J}_{k}\} \eqqcolon \mathcal{I}. $$ 
Here,  $\mathcal{J}_{1}, ..., \mathcal{J}_{k}$ lie in some large integer hash space such as $\{1,...,2^{32}\}$. Each $\mathcal{J}_i$ is constructed to correspond to a distinct circular chemical substructure detected within $\mathcal{M}$ (up to rare hash collisions that cause an identifier $\mathcal{J}_i$ to be associated with multiple substructures~\cite{rogers2010extended,gutlein2016filtered}). The set $\mathcal{I}$ can intuitively be thought of as the set of circular chemical substructures present in $\mathcal{M}$. The computational step
$$\mathcal{M} \mapsto \mathcal{I} \subseteq \{1,...,2^{32}\} $$
is referred to as \textit{substructure enumeration} by us and depends on two important hyperparameters: the maximal diameter, $D \in \{0,2,4,6,...\}$, up to which circular substructures should be detected; and the chosen list of atomic invariants,~$A$, used to distinguish the atoms in the input compound (such as the atomic number or whether the atom is part of a ring). The number of detected substructures, $k$, naturally varies with the input compound $\mathcal{M}$ the maximal substructure diameter $D$ and the selected atomic invariants $A$.

For further computational processing, for example by a machine-learning system, the set of substructure identifiers~$\mathcal{I}$ is commonly converted into a high-dimensional binary vector, $\mathcal{F} \in \{0,1\}^{L}$, via a simple hashing procedure. Let $$h : \{1, ..., 2^{32}\} \to \{1, ..., L\}$$
be a common (arbitrary) hash function that compresses the hash space $\{1, ..., 2^{32}\}$ into a much smaller space $\{1,...,L\}$. Then $\mathcal{F}$ is defined componentwise via:
$$\mathcal{F}_i = \left\{
\begin{array}{ll}
1 & \quad \exists \ \mathcal{J} \in \mathcal{I}: h(\mathcal{J}) = i , \\
0 & \quad \text{else}. \\
\end{array}
\right. $$
The computational step
$$\mathcal{I} \mapsto \mathcal{F} \in \{0,1\}^L $$
is referred to as \textit{hash-based folding} by us and depends on the chosen fingerprint dimension, $L \in \mathbb{N}$, as a hyperparameter. Common choices for $L$ include $1024$ or $2048$; and less often $512$ or $4096$. The binary vectorial components of $\mathcal{F}$ indicate (up to hash collisions in $\{1,...,L\}$) the existence of particular substructure identifiers in $\mathcal{I}$ which subsequently indicate (again, up to rare hash collisions~\cite{rogers2010extended,gutlein2016filtered}) the existence of particular circular chemical substructures in the input compound $\mathcal{M}$. The hashed feature vector $\mathcal{F}$ thus encodes valuable structural information about $\mathcal{M}$ and can for instance be fed into a machine-learning system for molecular property prediction.

\subsection*{Research Aims and Contributions}

The technical parallels between ECFPs and current message-passing GNNs~\cite{gilmer2017neural,duvenaud2015convolutional,xu2018powerful,kipf2016semi,yang2019analyzing,wu2020comprehensive, wieder2020compact,liu2019chemi} are striking. In both cases, a compound is first transformed into an (unordered) set-representation: for ECFPs, into a set of integer substructure identifiers, $\mathcal{I}$; for GNNs, into a set of initial and updated atom feature vectors\footnote{\footnotesize To be precise, one uses multisets (i.e.,~sets with counts) instead of classical sets for GNN-architectures to be able to distinguish identical atom feature vectors belonging to distinct atoms. Similarly, one would use multisets instead of sets when dealing with ECFPs with counts instead of binary ECFPs. However, in this work we focus on binary ECFPs owing to their more widespread use and conceptual simplicity.}. For both methods, a \textit{pooling operation} then plays the crucial role of reducing the given set-representation to a compound-wide feature vector. 

While considerable work has been done to investigate pooling methods for GNN-architectures~\cite{navarin2019universal,cangea2018towards,lee2019self,ranjan2020asap,ma2020path}, almost no analogous research exists on \textit{substructure pooling} for ECFPs or other structural fingerprints. Currently, hash-based folding is the default substructure-pooling technique for ECFPs to vectorise the set $\mathcal{I}$ and its use is ubiquitous in the molecular machine learning and cheminformatics literature~\cite{rogers2010extended,gutlein2016filtered,riniker2013open,duvenaud2015convolutional, webel2020revealing,zhong2023count,harada2020dual,ucak2022retrosynthetic,capecchi2020one,le2020neuraldecipher,shen2019molecular,tripp2024tanimoto}, although a few alternative hashing procedures have been explored by Probst and Reymond~\cite{probst2018probabilistic} for analog searches in big data settings. In spite of its widespread use, hash-based folding comes with a considerable downside: distinct substructural identifiers in $\mathcal{I}$ can be hashed to the same binary component of the output vector $\mathcal{F}$. Such ``bit collisions" necessarily occur when the fingerprint dimension $L$ is smaller than the number of circular substructures detected across a set of training compounds, which is almost always the case in standard settings. The ambiguities introduced by bit collisions into the fingerprint not only compromise its interpretability but also its predictive performance in machine-learning applications.

In this work, we describe a very simple and surprisingly effective alternative to hash-based folding for the pooling of ECFP substructures which we call \textit{Sort \& Slice}. In a nutshell, Sort \& Slice first sorts all circular substructures in the training set according to the respective number of training compounds in which they occur, and then ``slices" away all but the $L$ most frequent substructures which are subsequently used to generate an $L$-dimensional bit-collision-free binary fingerprint. The absence of bit collisions substantially improves the interpretability of Sort \& Slice ECFPs compared to their hashed counterparts, which regularly contain ambiguous components that correspond to multiple substructure identifiers. In spite of its simplicity, one can show that Sort \& Slice almost exclusively selects only the most informative substructural features from an entropic point of view~\cite{shannon1948mathematical,cover1991entropy}. The only other study we discovered in our literature search that explores a method similar to Sort \& Slice is the useful work of MacDougall~\cite{macdougall2022reduced}. Our study represents a more extensive and robust investigation of a closely related but further developed method that exhibits several additional stengths. 

In summary, we aim to provide the following contributions in this work:

\begin{itemize}
	
	\item We introduce a general mathematical definition of substructure pooling for structural fingerprints that subsumes hash-based folding, Sort \& Slice, and a large number of other potential techniques under a single theoretical framework.
	
	\item We give a technical description of Sort \& Slice as an easy-to-implement and bit-collision-free alternative to hash-based folding for ECFPs.
	
	\item We show via a series of rigorous computational experiments that Sort \& Slice robustly outperforms standard hash-based folding~\cite{rogers2010extended} across distinct ECFP-based molecular property prediction tasks, data splitting techniques, machine learning models, fingerprint diameters $D$, fingerprint dimensions $L$, and atomic invariants $A$; and that frequently the performance gains associated with Sort \& Slice are surprisingly large. We also show that Sort \& Slice outperforms two advanced substructure-selection schemes: filtered fingerprints by Gütlein and Kramer~\cite{gutlein2016filtered} and mutual-information maximisation~(MIM)~\cite{shannon1948mathematical,cover1991entropy}. 
	
	\item We recommend that due to its technical simplicity, improved interpretability and superior predictive performance, Sort \& Slice should canonically replace hash-based folding as the default substructure pooling method to vectorise ECFPs for supervised molecular machine learning.
	
\end{itemize}

\section*{Methodology}

\subsection*{Substructure Pooling: Definition and Setting}

We start by introducing a general mathematical definition of substructure pooling that can be used in combination with ECFPs or other structural fingerprints.
\begin{definition}[Substructure Pooling] \label{def: substructure_pooling}
	Let
	$$\mathfrak{J} = \{\mathcal{J}_{1}, ..., \mathcal{J}_{m}\} $$
	be a finite (but potentially very large) set of $m$ chemical substructures. The substructures $\mathcal{J}_1, ..., \mathcal{J}_m$ could be specified via hashed integer identifiers, SMILES strings, molecular graphs, or any other computational representation.
	Now, let the set of all possible subsets of $\mathfrak{J}$ (i.e.,~its power set) be denoted by
	$$ P(\mathfrak{J}) \coloneqq \{\mathcal{I} \ \vert \ \mathcal{I} \subseteq  \mathfrak{J} \}. $$
	A substructure-pooling method of dimension $L \in \mathbb{N}$ for the chemical substructures in $\mathfrak{J}$ is a set function
	$$\Psi :  P(\mathfrak{J}) \to \mathbb{R}^L,$$
	i.e.,~a function that maps arbitrary subsets of $\mathfrak{J}$ to $L$-dimensional real-valued vectors.
\end{definition}

Note that the fact that $\Psi$ is a set function implicitly presupposes that it is permutation-invariant, i.e., that its output does not depend on any arbitrary ordering of the substructures in the input set. One can further imagine $\Psi$ to not necessarily be a fixed function; it could also be a trainable deep network.

Substructure pooling naturally appears in supervised molecular machine learning during the vectorisation of structural fingerprints. Consider a supervised molecular property prediction task specified by a training set $\mathfrak{T}$ of $n$ unique compounds
$$\mathfrak{T} = \{\mathcal{M}_1, ..., \mathcal{M}_n\} $$
and an associated function
$$f_{\text{labels}} : \mathfrak{T} \to \mathbb{R}$$
that assigns regression or classification labels to the training set.
The training compounds $\mathcal{M}_1, ..., \mathcal{M}_n$ could for instance be represented as SMILES strings or molecular graphs and are assumed to be elements of some larger space of chemical compounds $\mathfrak{M}$ with
$$\mathfrak{M} \supset \{\mathcal{M}_1, ..., \mathcal{M}_n\} = \mathfrak{T}.$$ Now let
$$\mathfrak{J} = \{\mathcal{J}_{1}, ..., \mathcal{J}_{m}\} $$
be a set representing $m$ chemical substructures of interest. $\mathfrak{J}$ could for instance be the set of hashed integer identifiers of all possible ECFP substructures with chosen atomic invariants $A$ and maximal diameter $D$ or a set of $166$ functional groups represented via their respective SMARTS strings~\cite{sayle19971st} as with MACCS fingerprints~\cite{durant2002reoptimization}. Now let
$$\varphi : \mathfrak{M} \to P(\mathfrak{J})$$
describe the substructure-enumeration algorithm of a structural fingerprint. Then $\varphi$ maps a compound $\mathcal{M} \in \mathfrak{M}$ to the subset of substructures in $\mathfrak{J}$ that are contained in $\mathcal{M}$. Via $\varphi$ one can transform each training compound $\mathcal{M}_i$ into a set-representation, $\mathcal{I}_i$, consisting of $k_i$ substructures
$$\varphi(\mathcal{M}_i) = \{\mathcal{J}_{i,1}, ..., \mathcal{J}_{i, k_i} \} \eqqcolon \mathcal{I}_i \subseteq \mathfrak{J}. $$
At this stage, one requires a substructure-pooling method
$$\Psi :  P(\mathfrak{J}) \to \mathbb{R}^L$$
to transform each molecular set-representation $\mathcal{I}_i$ into a real-valued vector 
$$\Psi(\mathcal{I}_i) \eqqcolon \mathcal{F}_i \in \mathbb{R}^L.$$ 
The vector-representations $\mathcal{F}_1,...,\mathcal{F}_n$ can then be fed into a standard machine learning model which can be trained to predict the labels specified by~$f_{\text{labels}}$. Notice that the substructure-pooling operator $\Psi$ can be constructed leveraging knowledge from $\mathfrak{T}$ and $f_{\text{labels}}$.

The problem of substructure pooling can easily be translated into a mathematical problem that resembles GNN pooling if one employs a substructure embedding of some dimension, $w$:
$$\gamma : \mathfrak{J} \to \mathbb{R}^{w}. $$
Using $\gamma$, one can transform substructure sets into vector sets:
$$\mathfrak{J} \supseteq \{\mathcal{J}_{1}, ..., \mathcal{J}_{k}\} \mapsto \{\gamma(\mathcal{J}_{1}), ..., \gamma(\mathcal{J}_{k})\} \subset \mathbb{R}^w. $$ 
Many GNN-pooling methods, such as summation, averaging, max-pooling, or the differentiable operator proposed by Navarin et al.~\cite{navarin2019universal}, correspond to simple (graph-topology-independent) set functions that map vector sets to single vectors. All such functions can be repurposed to vectorise sets of embedded substructures. For example, by composing the sum operator $\Sigma$ with $\gamma$ we immediately gain a substructure-pooling method whose output dimension~$L$ is equal to the embedding-dimension~$w$:
\begin{equation*}
\Psi :  P(\mathfrak{J}) \to \mathbb{R}^w, \\
\Psi(\{\mathcal{J}_{1}, ..., \mathcal{J}_{k}\}) = \sum_{i = 1}^k \gamma(\mathcal{J}_{i}).
\end{equation*}
Perhaps the simplest possible substructure embedding is given via \textit{one-hot encoding}. 
\begin{definition}[One-Hot Encoding] \label{def: one_hot_enc}
	Let $u^w_{i} \in \mathbb{R}^w$ be the $w$-dimensional unit vector with $1$ in its $i$-th component and $0$ everywhere else. Furthermore, let 
	$$s: \{\mathcal{J}_{1}, ..., \mathcal{J}_{w}\} \to \{1,...,w\} $$
	be a bijective function that imposes a strict total order on a set of $w$ substructures $\{\mathcal{J}_{1}, ..., \mathcal{J}_{w}\}$ by assigning a unique rank to each substructure. Then we define the one-hot encoding associated with $s$ as
	$$\gamma_s : \{\mathcal{J}_{1}, ..., \mathcal{J}_{w}\} \to \mathbb{R}^{w}, \quad \gamma_s(\mathcal{J}) = u^w_{s(\mathcal{J})}. $$
\end{definition}
If the embedding dimension $w$ is equal to the total number of considered substructures $\vert \mathfrak{J} \vert = m$, it can be extremely large and additional dimensionality-reduction techniques might be required.

Finally, note that while in this work we focus on substructure pooling for binary structural fingerprints due to their simplicity and widespread use, it would be possible to further generalise Definition~\ref{def: substructure_pooling} to fingerprints with substructure counts by extending the domain of $\Psi$ to multisets (i.e.,~sets with counts) of substructures instead of classical sets.

\subsection*{Investigated Substructure-Pooling Techniques for ECFPs}

We now go on to use the theoretical framework introduced in the previous section to describe four distinct substructure-pooling techniques for ECFPs that we investigate in our study. From here on, let 
$$\mathfrak{J} = \{\mathcal{J}_{1}, ..., \mathcal{J}_{m}\} \subseteq \{1,...,2^{32}\}$$ 
be the set of hashed integer identifiers of \textit{all} possible ECFP substructures based on a predefined list of atomic invariants, $A$, and a fixed maximal diameter, $D \in \{0,2,4,6,...\}$. Further, let
$$\mathfrak{T} = \{\mathcal{M}_1, ..., \mathcal{M}_n\} \supset \mathfrak{M} $$
be a training set of $n$ compounds (for example represented as SMILES strings) drawn from a larger space of chemical compounds, $\mathfrak{M}$, let
$$f_{\text{labels}} : \mathfrak{T} \to \mathbb{R}$$
be a function that assigns regression or classification labels to the training set, and let
$$\varphi : \mathfrak{M} \to P(\mathfrak{J}), \quad \varphi(\mathcal{M}) = \{ \mathcal{J}_1,...,\mathcal{J}_k \} \subseteq \mathfrak{J},$$
be the ECFP-substructure enumeration algorithm that maps a compound $\mathcal{M}~\in~\mathfrak{M}$ to the set of circular substructures $\{ \mathcal{J}_1,...,\mathcal{J}_k \}$ in $\mathfrak{J}$ that are chemically contained in $\mathcal{M}$. Moreover, for each substructure identifier $\mathcal{J} \in \mathfrak{J}$, 
we define its associated binary substructural feature in the training set via
$$ f_{\mathcal{J}} :  \mathfrak{T} \to \{0,1\}, \quad f_{\mathcal{J}}(\mathcal{M}) = \left\{
\begin{array}{ll}
1 & \quad \mathcal{J} \in \varphi(\mathcal{M}), \\
0 & \quad \text{else}, \\
\end{array}
\right.  $$
and its \textit{support} as the set of all training compounds that contain $\mathcal{J}$,
$$ \text{supp}_{\mathfrak{T}}(\mathcal{J}) \coloneqq \{ \mathcal{M} \in \mathfrak{T} \ \vert \  f_{\mathcal{J}}(\mathcal{M}) = 1 \}.$$
Finally, let the set
$$\mathfrak{J}_{\mathfrak{T}} \coloneqq \bigcup_{\mathcal{M} \in \mathfrak{T}} \varphi(\mathcal{M}) $$
with cardinality $\vert~\mathfrak{J}_{\mathfrak{T}}~\vert~\eqqcolon~m_{\mathfrak{T}}$ represent all \textit{training substructures}, i.e.,~all substructure identifiers in $\mathfrak{J}$ that form part of at least one of the $n$ training compounds.

\subsubsection*{Hash-Based Folding}

The default way for ECFP-based substructure pooling~\cite{rogers2010extended} is by making use of a common (arbitrary) hash function
$$h : \{1, ..., 2^{32}\} \to \{1, ..., L\}.$$
Formally, one can employ $h$ to define a substructure-pooling method
$$\Psi :  P(\mathfrak{J}) \to \mathbb{R}^L$$
by setting its $i$-th output component to
\begin{equation*}
\begin{gathered}
\Psi(\{\mathcal{J}_1,...,\mathcal{J}_k\})_i = \\[1ex]
\left\{
\begin{array}{ll}
1 & \quad \exists \ \mathcal{J} \in \{\mathcal{J}_1,...,\mathcal{J}_k\}: h(\mathcal{J}) = i , \\
0 & \quad \text{else}. \\
\end{array}
\right.
\end{gathered}
\end{equation*}
The composite map
$$\Psi \circ \varphi : \mathfrak{M} \to \mathbb{R}^L $$
transforms an input compound $\mathcal{M} \in \mathfrak{M}$ into an $L$-dimensional binary vector whose $i$-th component is $1$ if and only if (at least) one of the substructures in $\varphi(\mathcal{M}) = \{\mathcal{J}_1,...,\mathcal{J}_k\}$ gets hashed to the integer $i$. This substructure-pooling method is straightforward to describe and implement. However, hash collisions in $\{1,...,L\}$ necessarily start to degrade its interpretability and predictive performance if $L$ becomes smaller than $m_{\mathfrak{T}}$. Note that $\Psi$ is independent of $\mathfrak{T}$ and $f_{\text{labels}}$.

\subsubsection*{Sort \& Slice}
Let
$$c : \mathfrak{J}_{\mathfrak{T}} \to \{1,...,n\}, \quad c(\mathcal{J}) = \vert \text{supp}_{\mathfrak{T}}(\mathcal{J}) \vert, $$
be a function that assigns to every training substructure $\mathcal{J} \in \mathfrak{J}_{\mathfrak{T}}$ the number of training compounds in which it is contained. Then we use $c$ to define a strict total order $\prec$ on $\mathfrak{J}_{\mathfrak{T}}$. For $\mathcal{J}, \tilde{\mathcal{J}} \in \mathfrak{J}_{\mathfrak{T}}$, we write $\mathcal{J} \prec \tilde{\mathcal{J}}$ if
$$ c(\mathcal{J}) < c(\tilde{\mathcal{J}}) \ \text{or} \ [ c(\mathcal{J}) = c(\tilde{\mathcal{J}}) \ \text{and} \ \mathcal{J} < \tilde{\mathcal{J}} ].$$
The order $\prec$ sorts substructures according to their frequencies in the training set, whereby ties are broken using the (arbitrary) order defined by the integer identifiers themselves. 
Let
$$s : \mathfrak{J}_{\mathfrak{T}} \to \{1,...,m_{\mathfrak{T}}\} $$
be a bijective \textit{sorting function} that assigns the ranks determined by $\prec$, whereby rank $1$ is assigned to the most frequent substructure.
Now let
$$\gamma_s : \mathfrak{J} \to \mathbb{R}^{m_{\mathfrak{T}}}, \quad \gamma_s(\mathcal{J}) = \left\{
\begin{array}{ll}
u^{m_{\mathfrak{T}}}_{s(\mathcal{J})} & \quad \mathcal{J} \in \mathfrak{J}_{\mathfrak{T}}, \\
(0,...,0) & \quad \text{else}. \\
\end{array}
\right.  $$
be an embedding that assigns a one-hot encoding sorted by $s$~(see~\Cref{def: one_hot_enc}) to training substructures and a vector entirely composed of zeros to substructures that do not appear in the training set. Based on $m_{\mathfrak{T}}$ and the desired fingerprint length $L$, we further define a \textit{slicing function}
\begin{equation*}
\begin{gathered}
\eta_{m_{\mathfrak{T}},L} : \mathbb{R}^{m_{\mathfrak{T}}} \to \mathbb{R}^L, \\[1ex]
\eta_{m_{\mathfrak{T}},L}(v_1,...,v_{m_{\mathfrak{T}}}) = \left\{
\begin{array}{ll}
(v_1,...,v_L) & L \leq m_{\mathfrak{T}}, \\
(v_1,...,v_{m_{\mathfrak{T}}},0,...,0) & L > m_{\mathfrak{T}}. \\
\end{array}
\right. 
\end{gathered}
\end{equation*}
Then the \textit{Sort \& Slice} substructure-pooling operator
$$\Psi :  P(\mathfrak{J}) \to \mathbb{R}^L $$
is given by
$$\Psi(\{\mathcal{J}_1,...,\mathcal{J}_k\}) = \eta_{m_{\mathfrak{T}},L} \Big( \sum_{i = 1}^k \gamma_s(\mathcal{J}_i) \Big).$$
The summation
$$\sum_{i = 1}^k \gamma_s(\mathcal{J}_i) \in \{0,1\}^{m_{\mathfrak{T}}} $$
corresponds to a sorted binary vector, each of whose components indicates the existence of a particular training substructure within the input set. Substructures that occur in more training compounds appear earlier in the vector. In the usual case where $L \leq m_{\mathfrak{T}}$, the function $\eta_{m_{\mathfrak{T}},L}$ trims the dimensionality of the vector to $L$ by slicing away the less frequent substructures. In the unusual case where $L$ is specifically demanded to be larger than $m_{\mathfrak{T}}$, all training substructures are contained in the final representation and it is simply padded with additional zeros.

The Sort \& Slice operator $\Psi$ is dependent on $\mathfrak{T}$ but not on $f_{\text{labels}}$. It can be interpreted as a simple unsupervised feature selection technique. In simplified terms, the composite map
$$\Psi \circ \varphi : \mathfrak{M} \to \mathbb{R}^L $$
outputs a binary $L$-dimensional fingerprint that represents the presence or absence of the $L$ most frequent training substructures in the input compound. In particular, this fingerprint does not suffer from bit collisions; each vectorial component can be assigned to a unique ECFP-substructure identifier. This clarity comes at the cost of losing information contained in the less frequent training substructures that are sliced away. 

Note however that in real-world chemical data sets the vast majority of ECFP substructures in the training set only occur in a few compounds, and almost all substructures occur in less than half of all compounds. For instance, the well-known lipophilicity data set from MoleculeNet~\cite{wu2018moleculenet} encompasses $n = 4200$ compounds which together lead to the enumeration of $m_{\mathfrak{T}} = 16903$ unique ECFP substructures with a maximal diameter of $D = 4$. Of these $16903$ substructures, $52.4$~\% occur only in a single compound, $87.5$~\% occur in $10$ or fewer compounds, $98.3$~\% occur in $100$ or fewer compounds, and almost all~($99.92$~\%) occur in fewer than half of all compounds. This frequency distribution is typical and highly stable across many chemical data sets. In a machine-learning context, slicing away infrequent training substructures should thus lead to very little information loss as it corresponds to the removal of almost-constant features. 

More specifically, Sort \& Slice has a natural interpretation in the context of information theory: The empirical information content~(i.e.,~Shannon entropy~\cite{shannon1948mathematical}) of a binary feature column $(f_{\mathcal{J}}(\mathcal{M}_i))_{i=1}^n$ peaks if the substructure $\mathcal{J}$ is contained in exactly half of all training compounds. If no substructure occurs in more than half of all training compounds, which is almost perfectly fulfilled for common data sets, then sorting substructures according to their training-set frequencies becomes equivalent to sorting columns of the training feature matrix according to their empirical information content. Sort \& Slice therefore automatically generates fingerprints that essentially only encompass the most informative training substructures from an entropic point of view, while being much simpler to describe, implement and interpret than an approach explicitly built on Shannon entropy. 

While theoretically Sort \& Slice might still lead to the inclusion of a tiny number of uninformative high-frequency training substructures, we decided against also slicing these away as this would have unnecessarily complicated our method for negligible expected performance gains. For example, out of the $16903$ ECFP substructures with maximal diameter $D=4$ in the MoleculeNet lipophilicity data set~\cite{wu2018moleculenet}, only $14$ substructures occur in more than half of all compounds, and only $3$ occur in $90$~\% or more.

Finally, note that we do not dare to claim that the principles behind Sort \& Slice were necessarily first discovered and utilised exclusively by us; due to its natural simplicity, variations of Sort \& Slice might have already been experimented with by other researchers in the past. However, to the best of our knowledge, this work represents the first thorough theoretical and computational investigation of Sort \& Slice in a peer-reviewed research paper. In particular, we are not aware of any other technique similar to Sort \& Slice that has been formally investigated, with the exception of one interesting method proposed by MacDougall~\cite{macdougall2022reduced}; however, this alternative algorithm does not appear to include a mechanism to break ties between substructures that are contained in the same number of training compounds. In~\cite{macdougall2022reduced}, infrequent substructures are thus sliced away in chunks which limits control over the fingerprint dimension $L$.

\subsubsection*{Filtering}

Gütlein and Kramer~\cite{gutlein2016filtered} conducted one of the few existing studies that systematically explores alternative substructure pooling strategies for ECFPs to circumvent the problem of bit collisions. They propose an advanced supervised substructure-selection scheme as a pooling method to construct what they refer to as \textit{filtered} fingerprints. Their method was originally published in \texttt{Java}; for this study, we reimplemented a version of it in \texttt{Python}. 

We assume that we are given a classification task with binary labels:
$$f_{\text{labels}} : \mathfrak{T} \to \{0,1\}. $$
If instead we are given a regression task with continuous labels, we binarise them by setting all labels below or above the label median to $0$ or $1$ respectively. 
Now let
$$ p_{\chi^2} : \mathfrak{J}_{\mathfrak{T}} \to [0,1],$$
be a function that assigns to each training substructure $\mathcal{J} \in \mathfrak{J}_{\mathfrak{T}}$ its $p$-value $p_{\chi^2}(\mathcal{J}) \in [0,1]$ in a statistical $\chi^2$ independence test~\cite{pearson1900chi2} based on the statistical sample $(f_{\text{labels}}(\mathcal{M}_i), f_{\mathcal{J}}(\mathcal{M}_i))_{i = 1}^n$ derived from the training compounds. The function $p_{\chi^2}$ defines a strict total order $\prec$ on $\mathfrak{J}_{\mathfrak{T}}$. For $\mathcal{J}, \tilde{\mathcal{J}} \in \mathfrak{J}_{\mathfrak{T}}$, we say $\mathcal{J} \prec \tilde{\mathcal{J}}$ if
$$ p_{\chi^2}(\mathcal{J}) > p_{\chi^2}(\tilde{\mathcal{J}}) \ \text{or} \ [ p_{\chi^2}(\mathcal{J}) = p_{\chi^2}(\tilde{\mathcal{J}}) \ \text{and} \ \mathcal{J} < \tilde{\mathcal{J}} ].$$
The smaller the $p$-value, the larger the substructure identifier according to $\prec$, whereby ties are broken using the (arbitrary) order defined by the integer identifiers themselves. 

Based on Gütlein and Kramers method, we go on to select a set of $L$ substructures $\mathfrak{J}_L$ from $\mathfrak{J}_{\mathfrak{T}}$ using the following strategy:
\begin{itemize}
	
	\item \textbf{Initialisation:} $\mathfrak{J}_L \coloneqq \mathfrak{J}_{\mathfrak{T}}.$
	
	\item \textbf{Step 1:}. A substructure $\mathcal{J} \in \mathfrak{J}_L$ that fulfills $\vert \text{supp}_{\mathfrak{T}}(\mathcal{J}) \vert = 1$ is randomly chosen and removed. This is repeated until all substructures in $\mathfrak{J}_L$ appear in at least two training compounds or until $\vert \mathfrak{J}_L \vert = L$.
	
	\item \textbf{Step 2:} A substructure $\mathcal{J} \in \mathfrak{J}_L$ that is \textit{non-closed} is randomly chosen and removed. This is repeated until all remaining substructures in $\mathfrak{J}_L$ are closed or until $\vert \mathfrak{J}_L \vert = L$. A substructure $\mathcal{J} \in \mathfrak{J}_L$ is called \textit{non-closed} if there exists another substructure $\tilde{\mathcal{J}} \in \mathfrak{J}_L$ such that $\text{supp}_{\mathfrak{T}}(\mathcal{J}) = \text{supp}_{\mathfrak{T}}(\tilde{\mathcal{J}})$ and $\mathcal{J}$ contains a proper subgraph that is isomorphic to $\tilde{\mathcal{J}}$.
	
	\item \textbf{Step 3:} The lowest-ranking element in $\mathfrak{J}_L$ with respect to the order $\prec$ is chosen and removed. This is repeated until $\vert \mathfrak{J}_L \vert = L$.
\end{itemize}
Step $1$ removes uninformative substructures that only occur in a single compound. Step $2$ represents a graph-theoretic attempt to reduce feature redundancy via the removal of substructures that contain smaller substructures that match the same set of training compounds. Finally, Step $3$ selects the $L$ remaining substructural features that show the strongest statistical dependence on the training label as quantified by a $\chi^2$-test. 

Using $\mathfrak{J}_L$ and an arbitrary bijective sorting function
$$s : \mathfrak{J}_L \to \{1,...,L\}$$
one can construct an embedding that generates an $L$-dimensional one-hot encoding for the selected substructures:
$$\gamma_s : \mathfrak{J} \to \mathbb{R}^{L}, \quad \gamma_s(\mathcal{J}) = \left\{
\begin{array}{ll}
u^L_{s(\mathcal{J})} & \quad \mathcal{J} \in \mathfrak{J}_L, \\
(0,...,0) & \quad \text{else}. \\
\end{array}
\right.  $$
Substructure pooling via filtering is then given by
$$\Psi :  P(\mathfrak{J}) \to \mathbb{R}^L, \quad \Psi(\{\mathcal{J}_1,...,\mathcal{J}_k\}) = \sum_{i = 1}^k \gamma_s(\mathcal{J}_i).$$
The composite map
$$\Psi \circ \varphi : \mathfrak{M} \to \mathbb{R}^L $$
transforms input compounds into binary fingerprints free of bit collisions that only encompass the selected substructures in~$\mathfrak{J}_L$. $\Psi$ depends on both $\mathfrak{T}$ and $f_{\text{labels}}$.

Note that the above algorithm for the construction of $\mathfrak{J}_L$ implicitly assumes the usual case where $L \leq m_{\mathfrak{T}}$. If for some reason $L$ is demanded to be larger than $m_{\mathfrak{T}}$, then we simply set $\mathfrak{J}_L \coloneqq \mathfrak{J}_{\mathfrak{T}}$ and pad the fingerprints generated by $\Psi$ with additional zeros up to dimension $L$.

\subsubsection*{Mutual Information Maximisation}

Substructure pooling via mutual-information maximisation~(MIM) is performed analogously to filtering, the only difference lying in the set of selected substructures $\mathfrak{J}_L$. We again assume that we are either given a binary classification task or a regression task that has been binarised around its median:
$$f_{\text{labels}} : \mathfrak{T} \to \{0,1\}. $$ 
Using the statistical sample $(f_{\text{labels}}(\mathcal{M}_i), f_{\mathcal{J}}(\mathcal{M}_i))_{i = 1}^n$ of binary training variables, one can compute the empirical \textit{mutual information} 
$$ \hat{I} : \mathfrak{J}_{\mathfrak{T}} \to [0, \infty)$$ 
between the training label and the feature associated with a training substructure $\mathcal{J} \in \mathfrak{J}_{\mathfrak{T}}$ using simple plug-in entropy estimators~\cite{shannon1948mathematical,cover1991entropy,zhang2011normal}.
$\hat{I}(\mathcal{J})$ is a nonnegative, symmetric and nonlinear measure of statistical dependence between $(f_{\text{labels}}(\mathcal{M}_i))_{i = 1}^n$ and $(f_{\mathcal{J}}(\mathcal{M}_i))_{i = 1}^n$. The larger $\hat{I}(\mathcal{J})$, the more information the presence of substructure $\mathcal{J}$ in a training compound conveys about its label and \textit{vice versa}. The function $\hat{I}$ defines a strict total order $\prec$ on $\mathfrak{J}_{\mathfrak{T}}$. For $\mathcal{J}, \tilde{\mathcal{J}} \in \mathfrak{J}_{\mathfrak{T}}$, we write $\mathcal{J} \prec \tilde{\mathcal{J}}$ if
$$ \hat{I}(\mathcal{J}) < \hat{I}(\tilde{\mathcal{J}}) \ \text{or} \ [\hat{I}(\mathcal{J}) = \hat{I}(\tilde{\mathcal{J}}) \ \text{and} \ \mathcal{J} < \tilde{\mathcal{J}} ].$$

We use the order $\prec$ to select $L$ substructures $\mathfrak{J}_L$ from $\mathfrak{J}_{\mathfrak{T}}$ via the following strategy (assuming $L \leq m_{\mathfrak{T}}$):
\begin{itemize}
	
	\item \textbf{Initialisation:} $\mathfrak{J}_L \coloneqq \mathfrak{J}_{\mathfrak{T}}.$
	
	\item \textbf{Step 1:} If for two substructures $\mathcal{J}, \tilde{\mathcal{J}} \in \mathfrak{J}_L$ it holds that $\text{supp}_{\mathfrak{T}}(\mathcal{J}) = \text{supp}_{\mathfrak{T}}(\tilde{\mathcal{J}})$, then either $\mathcal{J}$ or $\mathcal{\tilde{J}}$ is chosen uniformly at random and removed from $\mathfrak{J}_L$. This is repeated until no two substructures have the same support or until $\vert \mathfrak{J}_L \vert = L$.
	
	\item \textbf{Step 2:} The smallest element of $\mathfrak{J}_L$ with respect to the order $\prec$ is chosen and removed. This is repeated until $\vert \mathfrak{J}_L \vert = L$.
	
\end{itemize}
Step $1$ aims to reduce feature redundancy via the removal of identical substructural features from the training set. Step $2$ subsequently selects the $L$ training substructures that exhibit the highest mutual information with the training label. MIM depends on both $\mathfrak{T}$ and $f_{\text{labels}}$ and can thus be categorised as a supervised feature selection scheme. As in the case of filtering, in the unusual case where $L$ is demanded to be larger than $m_{\mathfrak{T}}$,
we simply set $\mathfrak{J}_L \coloneqq \mathfrak{J}_{\mathfrak{T}}$ and pad the resulting vectorial fingerprint with additional zeros up to dimension $L$.

Instead of MIM, we initially attempted to use \textit{conditional} MIM, a more advanced information-theoretic technique that iteratively selects features that maximise the mutual information with the training label \textit{conditional} on the information contained in any feature already picked. However, we observed that even the fast implementation of conditional MIM described by Fleuret~\cite{fleuret2004fast} was computationally slow to select thousands of substructures out or an even larger substructure pool. This made the generation of vectorial ECFPs with usual lengths such as $L = 1024$ or $L = 2048$ bits impractical. We thus decided to instead investigate classical MIM, which represents a natural simplification of conditional MIM that remains computationally feasible even in very high feature dimensions.

\subsection*{Categories of Substructure-Pooling Techniques}

\begin{figure*}[h!]

	\caption{Schematic overview of the vectorisation of ECFPs via the four investigated substructure-pooling techniques which in general lead to four different final representations. As before, $\mathfrak{T} = \{\mathcal{M}_1, ..., \mathcal{M}_n\}$ represents a given set of $n$ training compounds and $f_{\text{labels}} : \mathfrak{T} \to \mathbb{R}$ an associated labelling function that assigns regression or classification labels to the training set. }
	\vspace{12pt}

	\includegraphics[width=1.83\linewidth]{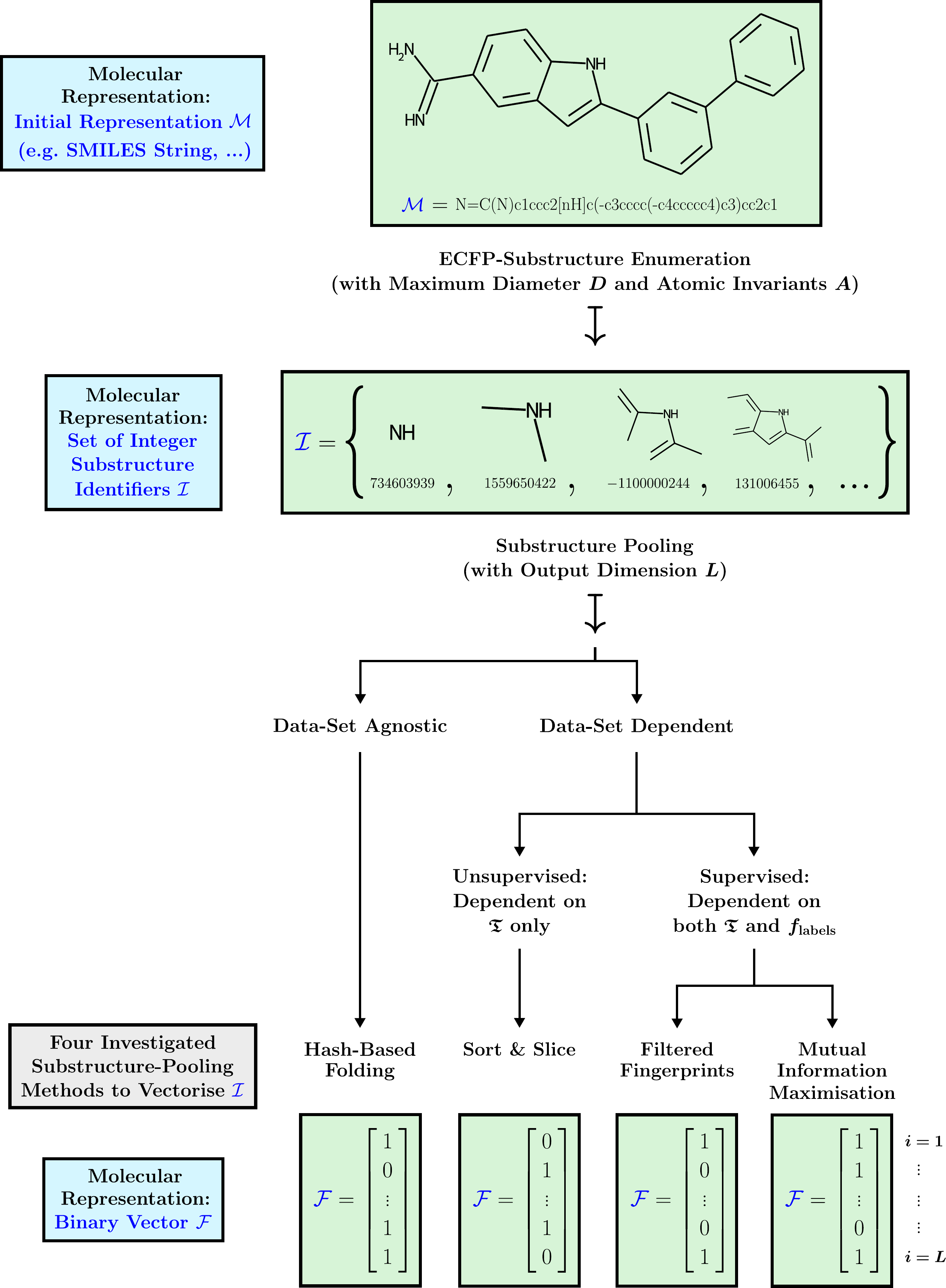}
	\label{fig:substructurepoolingmethodsforecfps}
\end{figure*}

As has been shown, substructure pooling as outlined in~\Cref{def: substructure_pooling} encompasses a large number of potential techniques. In this study, we investigate one data-set agnostic substructure-pooling procedure (classical hash-based folding) and three data-set dependent techniques~(Sort \& Slice, filtering, MIM). Our data-set dependent techniques are all based on either supervised or unsupervised feature selection. A graphical overview of the investigated methods can be found in~\Cref{fig:substructurepoolingmethodsforecfps}.

Note that there exist many more possibilities for substructure pooling, including techniques that are neither based on hashing nor on substructure selection. However, to the best of our knowledge, such methods remain entirely unexplored. For instance, an interesting avenue for future research might be the investigation of \textit{differentiable} substructure-pooling operators
$$\Psi_{\theta} :  P(\mathfrak{J}) \to \mathbb{R}^L$$ 
of the form 
$$\Psi_{\theta}(\{\mathcal{J}_1,...,\mathcal{J}_k\}) = \Phi_{\theta}(\{\gamma(\mathcal{J}_{1}), ..., \gamma(\mathcal{J}_{k})\}) $$
whereby
$$\gamma : \mathfrak{J} \to \mathbb{R}^{w} $$
could be a chemically meaningful substructure embedding and 
$$\Phi_{\theta} : \{ A \subset \mathbb{R}^w \ \vert \ A \text{ is finite}	\} \to \mathbb{R}^L $$
could be a trainable deep learning architecture designed to operate on finite sets of real-valued vectors~\cite{zaheer2017deep}.

\subsection*{Experimental Setting}

We computationally evaluated the relative performance of traditional hash-based folding, Sort \& Slice, filtering and MIM using five molecular property prediction data sets that were selected to cover a diverse set of chemical regression and classification tasks: the prediction of lipophilicity~\cite{wu2018moleculenet}, aqueous solubility~\cite{sorkun2019aqsoldb}, mutagenicity~\cite{hansen2009benchmark}, and binding affinity for SARS-CoV-2 main protease (experimentally measured via a fluorescence-based assay)~\cite{covid2020open}. We also included a highly imbalanced LIT-PCBA virtual screening data set for estrogen receptor $\alpha$ antagonism~\cite{tran2020lit}, whereby we used the full raw data rather than the preprocessed unbiased splits provided in~\cite{tran2020lit}. All experiments were implemented in \texttt{Python}.

The data sets consisted of SMILES strings which we algorithmically standardised and desalted using the ChEMBL structure pipeline~\cite{bento2020open}. This step also removed solvents and isotopic information. SMILES strings that generated error messages upon being turned into an \texttt{RDKit} mol object were deleted. Afterwards, if two standardised SMILES strings were found to be duplicates, one was deleted uniformly at random along with its training label. The only two data sets that contained noteworthy numbers of duplicates were the aqueous solubility and the LIT-PCBA data set; for these two cases, removing duplicates reduced the number of compounds from $9978$ to $9821$, and from $5045$ to $3921$ respectively. In the aqueous solubility data set, we further detected a few instances where SMILES strings appeared to encode several disconnected fragments; we also deleted such occurrences which reduced the number of compounds from $9821$ to the final size of $9335$. The five standardised and cleaned data sets are summarised in~\Cref{tab: data_sets_substruc_pool}.
\begin{table*}[!b]

	\caption{\centering Overview of the five molecular property prediction data sets used for our computational experiments.}
	\vspace{6pt}

	{\renewcommand{\arraystretch}{1.6}

		\begin{tabular}{V{2.3} p{6.5cm}| p{1.7cm} | p{1.9cm} | p{3.8cm} V{2.3} }

			\multicolumn{1}{l}{\textbf{Prediction Task}} 
			
			& \multicolumn{1}{l}{\textbf{Task Type}} 
			
			& \multicolumn{1}{l}{\textbf{Compounds}}
			
			& \multicolumn{1}{l}{\textbf{Source}} \\ \hlineB{2.3}

			Lipophilicity [logD] & Regression & $4200$  & MoleculeNet~\cite{wu2018moleculenet} \\ \hline 
			
			Aqueous Solubility [logS] & Regression & $9335$  & Sorkun et al.~\cite{sorkun2019aqsoldb} \\ \hline 
			
			SARS-CoV-2 Main Protease Binding Affinity [pIC\textsubscript{50}] & Regression & $1924$  & COVID Moonshot Project~\cite{covid2020open} \\ \hline 
			
			Ames Mutagenicity & Classification & $3496$ positives \newline $3009$ negatives & Hansen et al.~\cite{hansen2009benchmark} \\ \hline 
			
			Estrogen Receptor $\alpha$ Antagonism & Classification & $88$ positives \newline $3833$ negatives  & LIT-PCBA~\cite{tran2020lit} \\ \hlineB{2.3}

	\end{tabular}}

	\label{tab: data_sets_substruc_pool}

\end{table*}

As a data splitting strategy, we implemented $2$-fold cross validation repeated with $3$ random seeds for all data sets. Thus, each model was separately trained and tested $2 * 3 = 6$ times. We ran two distinct versions of our cross validation scheme for all experiments, based on either scaffold~\cite{bemis1996properties} or random data splitting (stratified random splitting for the imbalanced LIT-PCBA data set). Scaffold splitting guarantees that the scaffold of each training-set compound is distinct from the scaffold of each test-set compound. This creates a distributional shift between training and test set that leads to a more challenging prediction task. Performance results were recorded as the mean and standard deviation over the $6$ splits, using the mean absolute error~(MAE) for the three regression data sets, the area under the receiver operating characteristic curve~(AUROC) for the balanced mutagenicity classification data set, and the area under the precision recall curve~(AUPRC) for the imbalanced LIT-PCBA virtual screening data set.

As machine-learning models, we selected random forests~(RFs) and multilayer perceptrons~(MLPs). The RFs were implemented in scikit-learn~\cite{pedregosa2011scikit} and we chose the default hyperparameters for them with the exception of MaxFeatures for RF-regressors which was set to ``Sqrt" instead of $1.0$ to add randomness. The MLPs were implemented in \texttt{PyTorch}~\cite{paszke2019pytorch} and our MLP architecture consisted of five hidden layers, each with $512$ neurons, an additive bias vector, ReLU activations, batch normalisation~\cite{ioffe2015batch}, and a dropout rate~\cite{srivastava2014dropout} of $0.25$. For MLP training we employed an initial learning rate of $10^{-3}$, a learning rate decay factor of $0.98$ per epoch until a learning rate of $10^{-5}$ was reached, $250$ training epochs, a batch size of $64$, and AdamW optimisation~\cite{loshchilov2017decoupled} with a weight decay factor of $0.1$. For MLP-regressors we used identity output activations and a mean squared error loss; for MLP-classifiers we used sigmoidal output activations and a binary cross-entropy loss. All MLPs were trained on a single NVIDIA GeForce RTX 3060 GPU. 

ECFPs as implemented in \texttt{RDKit}~\cite{landrum2006rdkit} by default use a list of six standard atomic invariants, $A$, including for instance the atomic number and whether the atom is part of a ring. Alternatively, \texttt{RDKit} also provides a list of six binary pharmacophoric invariants, $\tilde{A}$ which more explicitly reflect the function an atom might play in pharmacological chemistry. These features for example include whether the atom is a halogen and whether it is a hydrogen bond acceptor. When using pharmacophoric invariants $\tilde{A}$ instead of standard invariants $A$, one speaks of functional-connectivity fingerprints~(FCFPs) instead of ECFPs. The selected diameter $D$ is usually appended to the fingerprint-abbreviation: for example, ECFP$4$ indicates a diameter of $D = 4$. Both ECFPs and FCFPs optionally allow for the stereochemical distinction between atoms with respect to tetrahedral R/S chirality; we decided to include this additional invariant in our experiments, in part motivated by the fundamental effects chirality is known to have in certain pharmacological settings~\cite{tokunaga2018understanding}; although it should be noted that chiral annotations may exhibit a certain level of noise in publicly curated data sets.

For each of $20$ modelling scenarios determined by selecting one of the five data sets, a data splitting technique (random or scaffold), and a machine-learning model (RF or MLP), we conducted a thorough and extensive investigation of the ECFP hyperparameter space. In each of the $20$ cases, we used $2$-fold cross validation repeated with $3$ random seeds to evaluate $96$ distinct binary vectorial fingerprints generated by exhaustively combining three maximal substructure diameters $D \in \{2, 4, 6\}$, two types of atomic invariants $A \in \{\text{standard ECFP, pharmacophoric FCFP}\}$, four fingerprint dimensions $L \in \{512, 1024, 2048, 4096\}$, and four substructure-pooling methods (hash-based folding, Sort \& Slice, filtering, or MIM). In total, the application of this robust combinatorial methodology resulted in the training of $20 * 96 * 2 * 3 = 11520$ separate machine-learning models, half of which were deep-learning models.

\section*{Results and Discussion} 

\begin{figure*}[h!]
	\centering
	\includegraphics[width=1.8\linewidth]{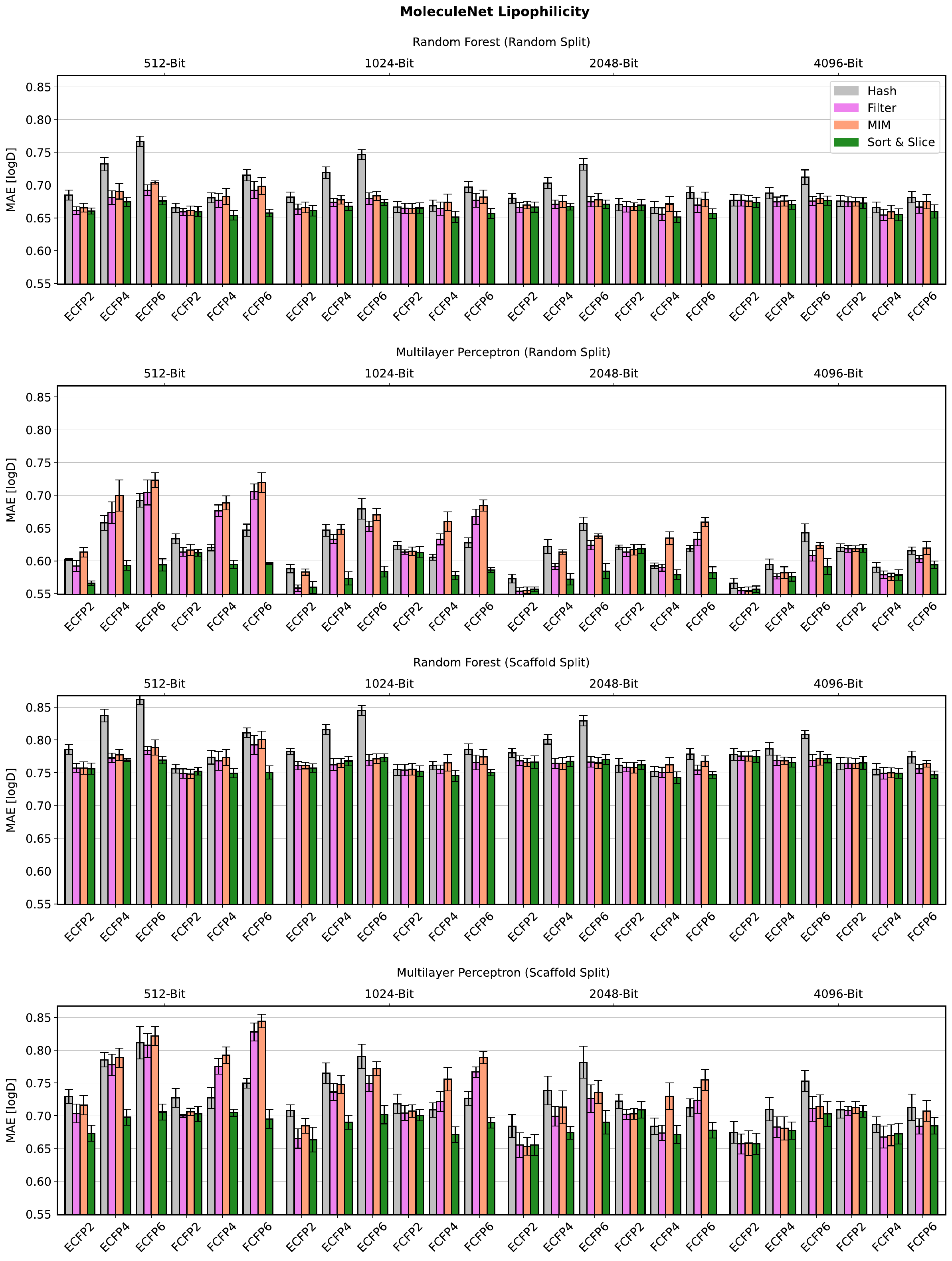}
	\caption{Predictive performance of the four investigated substructure-pooling methods (indicated by colours) for the lipophilicity regression data set using varying data splitting techniques, machine-learning models and ECFP hyperparameters. Each bar shows the average mean absolute error~(MAE) of the associated model across $2$-fold cross validation repeated with $3$ random seeds. The error bar length equals two standard deviations of the performance measured over the $2 * 3 = 6$ trained models.}
	\label{fig:moleculenet_lipophilicity}
\end{figure*}

\begin{figure*}[h!]
	\centering
	\includegraphics[width=1.62\linewidth]{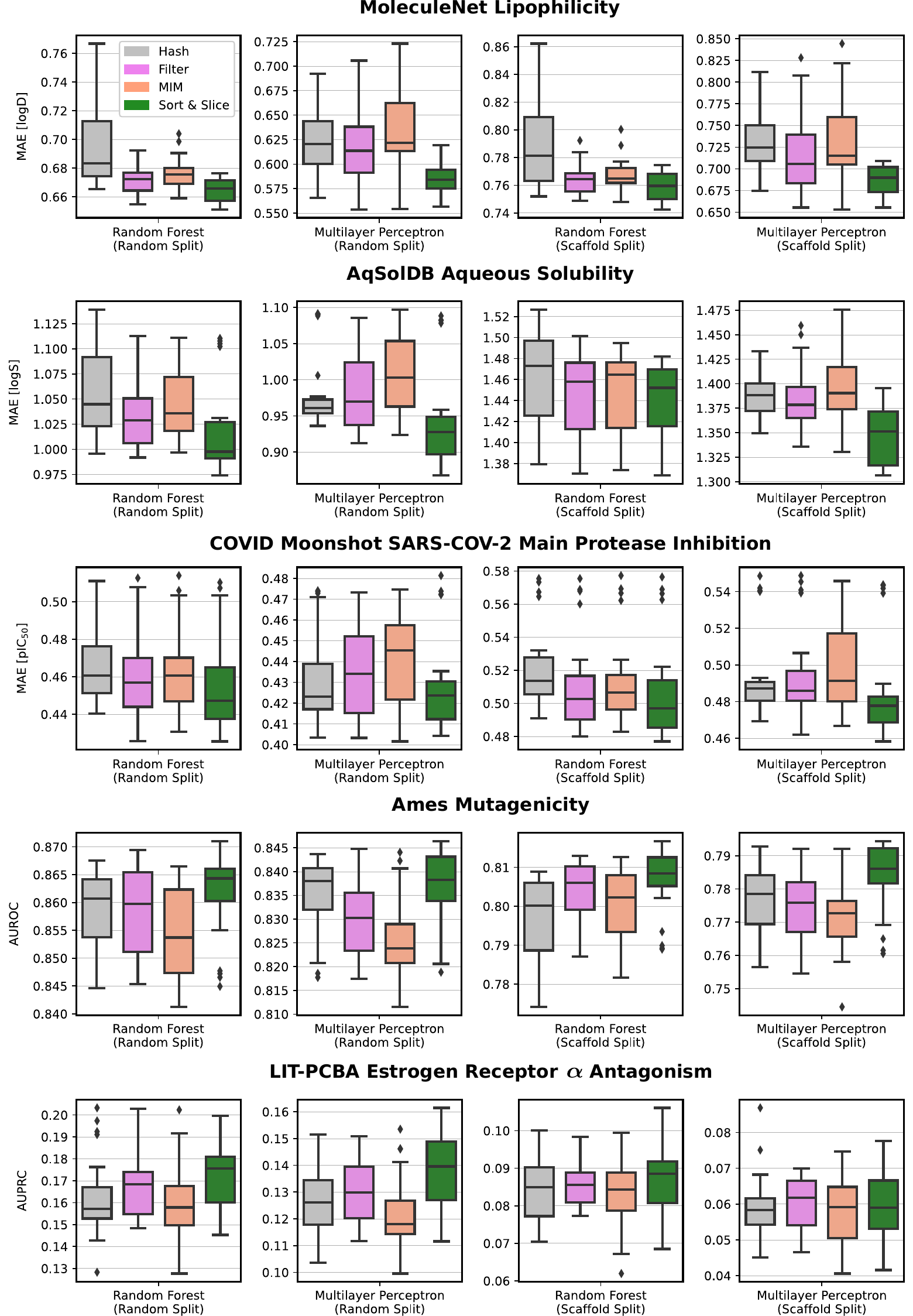}
	\caption{Boxplots visualising the predictive performance of the four investigated substructure-pooling methods (indicated by colours) for $20$ distinct modelling scenarios differing by data sets, data splitting techniques, and machine-learning models. Each boxplot summarises $24$ distinct performance results (each computed as an average over $2$-fold cross validation repeated with $3$ random seeds) generated by combining a respective substructure-pooling method with $24$ distinct ECFP-hyperparameter settings. The exhaustively explored ECFP-hyperparameter grid is given by three maximal substructure diameters $D \in \{2, 4, 6\}$, two lists of atomic invariants $A \in \{\text{standard ECFP, pharmacophoric FCFP}\}$, and four fingerprint dimensions $L \in \{512, 1024, 2048, 4096\}$. }
	\label{fig:boxplots_with_titles}
\end{figure*}

Detailed computational results for the lipophilicity prediction task are visualised in~\Cref{fig:moleculenet_lipophilicity}. A high-level summary of the results for all five data sets is depicted in~\Cref{fig:boxplots_with_titles}. Further detailed results for the remaining four data sets can be found in~\Cref{fig:aqsoldb_solubility,fig:postera_sars_cov_2_mpro,fig:ames_mutagenicity,fig:lit_pcba_esr_ant} in the appendix.


Overall, our numerical observations indicate the robust superiority of Sort \& Slice over hash-based folding. For instance, the Sort \& Slice version of the popular $1024$-bit ECFP$4$ surpasses the predictive performance of the hashed version in all but a few cases. For example, \Cref{fig:moleculenet_lipophilicity} reveals that simply replacing hash-based folding with Sort \& Slice  when using a $1024$-bit ECFP$4$ with an MLP for lipophilicity prediction with a random data split leads to a rather remarkable relative MAE-improvement of $11.37\%$. The results for mutagenicity prediction in~\Cref{fig:ames_mutagenicity} and for the virtual screening task in~\Cref{fig:lit_pcba_esr_ant} suggest that the advantage of Sort \& Slice over hash-based folding remains stable in both balanced and highly imbalanced classification scenarios. 

Parts of the detailed results in~\Cref{fig:moleculenet_lipophilicity,fig:aqsoldb_solubility,fig:postera_sars_cov_2_mpro,fig:ames_mutagenicity,fig:lit_pcba_esr_ant} suggest that the improvements achieved via Sort \& Slice over hash-based folding tend to become more pronounced as the fingerprint dimension $L$ decreases, the maximal substructure diameter $D$ increases and as standard atomic invariants are used instead of pharmacophoric invariants. Hashed ECFPs exhibit an increasing number of bit collisions as $L$ decreases relative to the number of detected substructures in the data set, which in turn tends to increase with $D$ and when switching from the abstract pharmacophoric to the more distinctive standard atomic invariants. An increase in bit collisions thus might degrade the performance of hashed ECFPs relative to the Sort \& Slice ECFP representations that by design are free of bit collisions.

A natural question to ask, however, is whether the predictive advantage of Sort \& Slice is merely a result of the general avoidance of bit collisions; or if and to what extent the particular unsupervised substructure-selection scheme underlying Sort \& Slice, i.e.,~the natural removal of low-variance features via the exclusion of substructures with low training-set frequency, contributes to the performance gain. Surprisingly,~\Cref{fig:boxplots_with_titles} shows that Sort \& Slice not only beats hash-based folding, but it also consistently outperforms filtering and MIM, two advanced supervised feature-selection schemes that just like Sort \& Slice are entirely free of bit collisions. More specifically, comparing MIM against hash-based folding delivers mixed results, with neither method appearing to be clearly superior to the other; filtering appears to mostly outperform both MIM and hash-based folding, but is in turn outperformed by Sort \& Slice.

These observations reveal two points: Firstly, the performance gains provided by Sort \& Slice over hash-based folding are not purely the result of avoiding bit collisions via substructure selection; but the specific substructure-selection strategy does indeed make an important difference for downstream predictive performance. Secondly, and perhaps remarkably, the extremely simple prevalence-based substructure-selection scheme implemented by Sort \& Slice outperforms the more technically advanced selection methods underlying filtering and MIM. This is in spite of the fact that, unlike MIM and filtering, Sort \& Slice is an unsupervised technique that does not utilise any information associated with the training label. It appears surprising that Sort \& Slice would beat technically sophisticated supervised feature selection methods such as filtering or MIM that select substructures using task-specific information. While the reasons for this are not obvious, it is conceivable that exploiting the training label when selecting substructures could potentially harm the generalisation abilities of a machine-learning system by contributing to its risk of overfitting to the training data (just like any other aspect of supervised model training). Another reason may be related to the natural frequency distribution of circular substructures in chemical data sets; the tendency for almost all ECFP substructures to appear in less than half of all training compounds causes the simple Sort \& Slice algorithm to essentially coincide with a more complex and potentially more powerful feature-selection strategy based on the maximisation of information entropy. Finally, unlike hash-based folding which is data-set agnostic, Sort \& Slice exploits structural information from the training compounds and could thus be interpreted as a simple and effective way to calibrate ECFPs to a specific region of chemical space.

\section*{Conclusions}

We have introduced a general mathematical framework for the vectorisation of structural fingerprints via a formal operation referred to as substructure pooling. For ECFPs, substructure pooling is a natural analogue to node feature vector pooling in message-passing GNN architectures. Unlike GNN pooling, ECFP-based substructure pooling is almost always performed using one particular method (hash-based folding); other substructure-pooling strategies for ECFPs remain largely unexplored. Our proposed substructure pooling framework encompasses hash-based folding, but also a broad spectrum of alternative methods that could be explored, including supervised and unsupervised substructure selection strategies. Trainable deep learning architectures operating on sets of chemically meaningful substructure embeddings also fit into the presented methodology; the investigation of such neural techniques, while out of the scope of this study, might form an interesting avenue for future research.

As part of our work, we have mathematically described and computationally evaluated \textit{Sort \& Slice}, a straightforward alternative to hash-based folding for the pooling of ECFP substructures that is very easy to implement and interpret. Sort \& Slice can be seen as a simple unsupervised feature-selection scheme that generates a binary vectorial fingerprint that is free of bit collisions and based only on the circular substructures that occur most frequently in the training compounds. Under realistic theoretical assumptions that are closely approximated by common chemical data sets, Sort \& Slice can be shown to select automatically only the most informative ECFP substructures from an information-theoretic perspective.

An extensive series of strictly conducted computational experiments indicates that Sort \& Slice tends to produce better (and sometimes substantially better) downstream performance than hash-based folding at ECFP-based molecular property prediction. Our numerical results suggest that the predictive advantage of Sort \& Slice is highly robust and exists across a diverse range of molecular regression and classification tasks, balanced and imbalanced classification settings, distinct data splitting techniques, machine-learning models and ECFP hyperparameters. Perhaps surprisingly, Sort \& Slice not only seems to outperform hash-based folding but also two technically sophisticated supervised substructure selection schemes~\cite{gutlein2016filtered,cover1991entropy}. This indicates that, in spite of its extreme simplicity, sorting ECFP substructures according to their relative prevalence in a given region of chemical space of interest and then discarding infrequent substructures is a (maybe unexpectedly) strong feature selection strategy. Based on the robust predictive advantage of Sort \& Slice, its technical simplicity, its ease of implementation, and its ability to improve fingerprint interpretability by avoiding bit collisions, we recommend that it should canonically replace hash-based folding as the default substructure-pooling technique to vectorise ECFPs for supervised molecular machine learning.

\section*{Appendix: Further Computational Results}

\begin{figure*}[h!]
	\centering
	\includegraphics[width=1.8\linewidth]{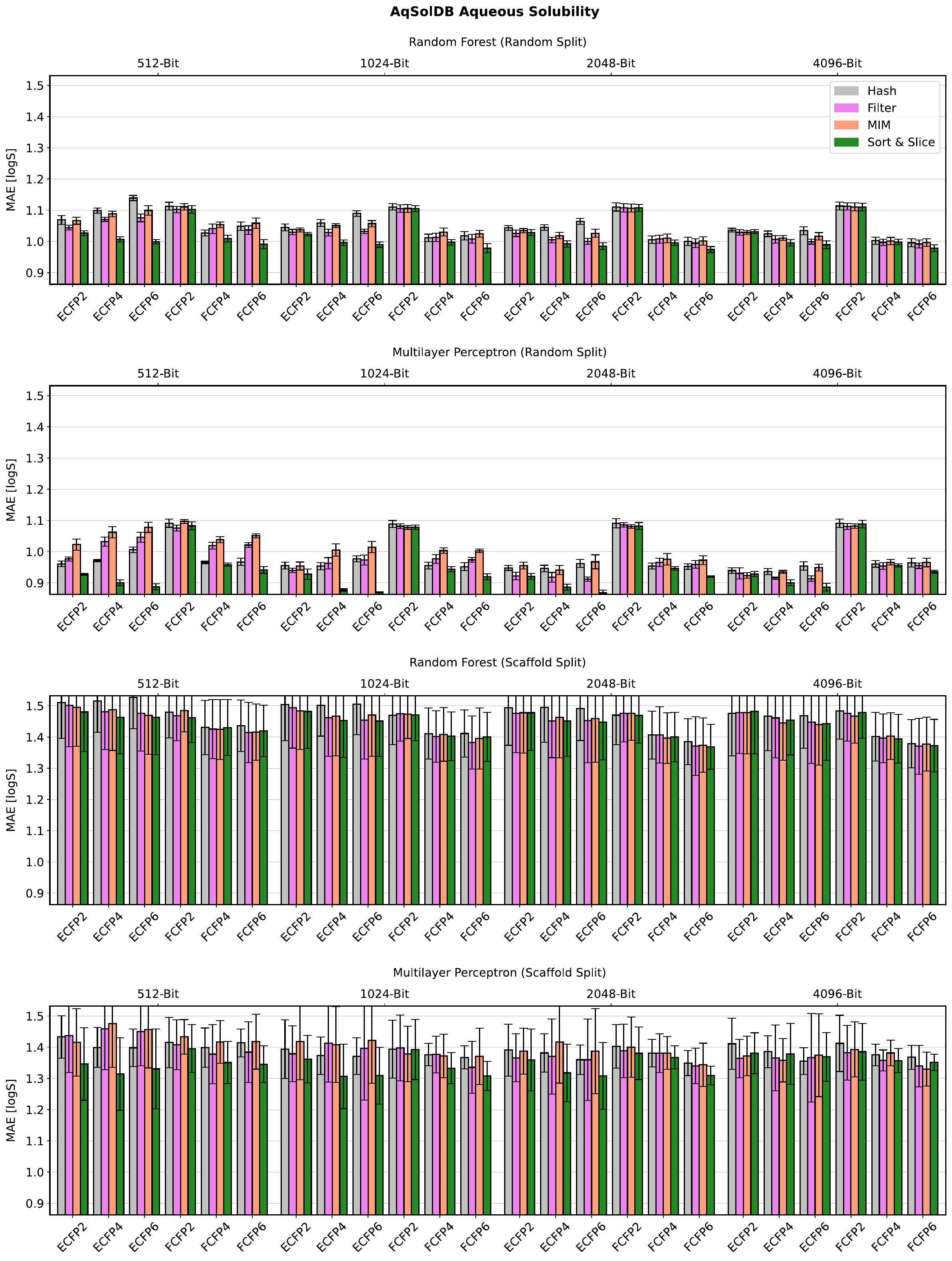}
	\caption{Predictive performance of the four investigated substructure-pooling methods (indicated by colours) for the aqueous solubility regression data set using varying data splitting techniques, machine-learning models and ECFP hyperparameters. Each bar shows the average mean absolute error~(MAE) of the associated model across $2$-fold cross validation repeated with $3$ random seeds. The error bar length equals two standard deviations of the performance measured over the $2 * 3 = 6$ trained models.}
	\label{fig:aqsoldb_solubility}
\end{figure*}

\begin{figure*}[h!]
	\centering
	\includegraphics[width=1.8\linewidth]{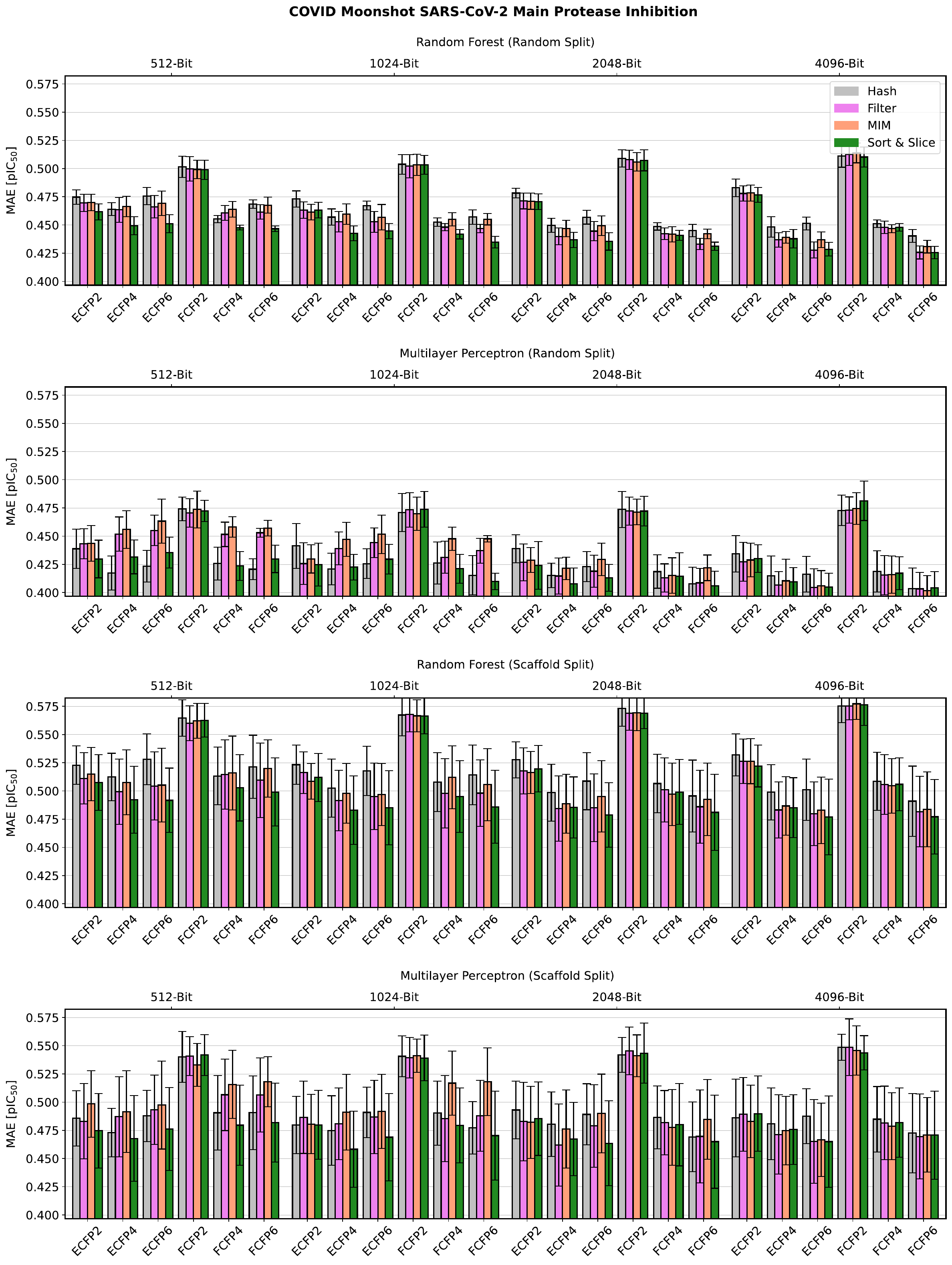}
	\caption{Predictive performance of the four investigated substructure-pooling methods (indicated by colours) for the SARS-CoV-2 main protease binding affinity regression data set using varying data splitting techniques, machine-learning models and ECFP hyperparameters. Each bar shows the average mean absolute error~(MAE) of the associated model across $2$-fold cross validation repeated with $3$ random seeds. The error bar length equals two standard deviations of the performance measured over the $2 * 3 = 6$ trained models.}
	\label{fig:postera_sars_cov_2_mpro}
\end{figure*}

\begin{figure*}[h!]
	\centering
	\includegraphics[width=1.8\linewidth]{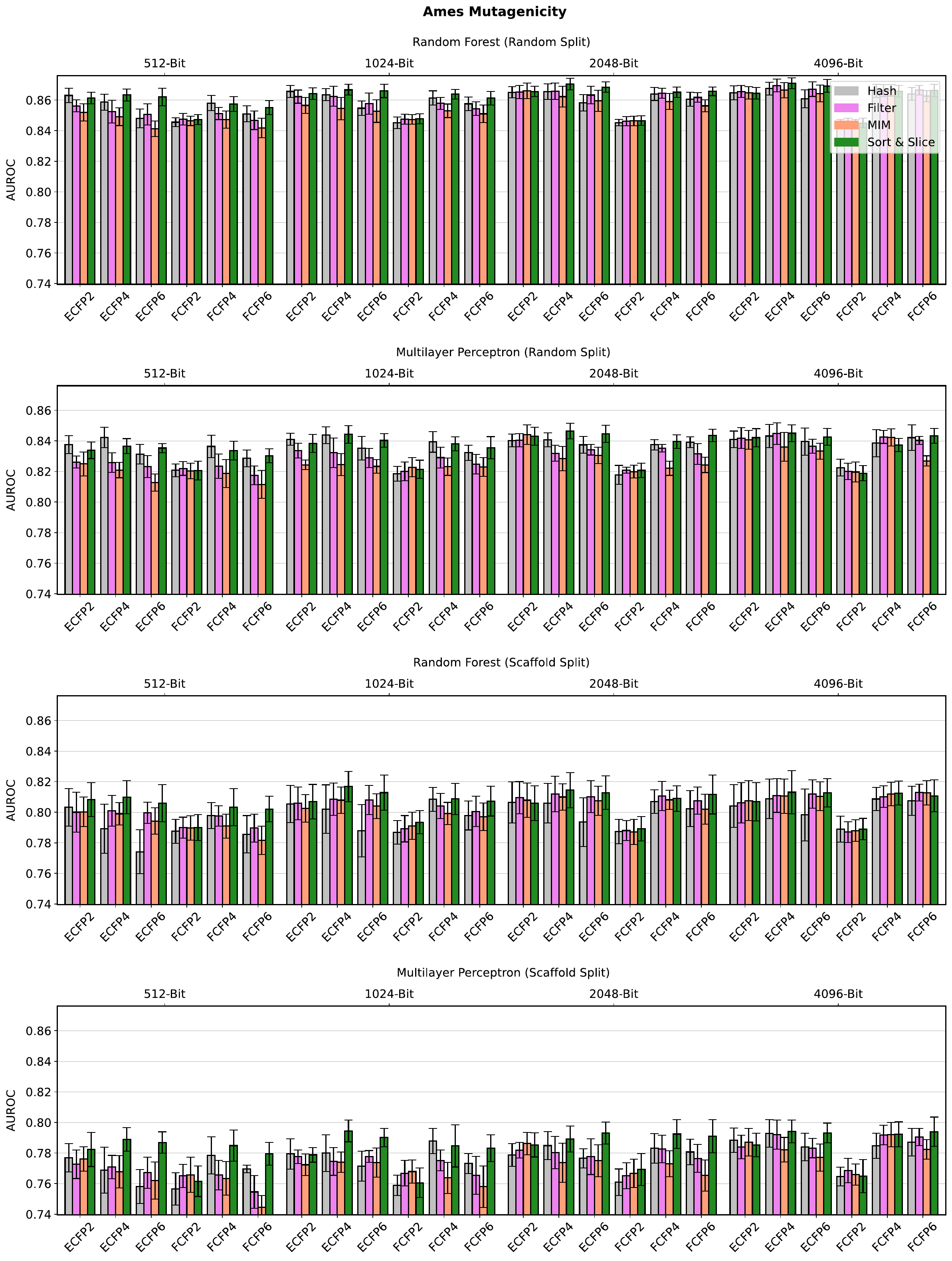}
	\caption{Predictive performance of the four investigated substructure-pooling methods (indicated by colours) for the balanced mutagenicity classification data set using varying data splitting techniques, machine-learning models and ECFP hyperparameters. Each bar shows the average average area under the receiver operating characteristic curve~(AUROC) of the associated model across $2$-fold cross validation repeated with $3$ random seeds. The error bar length equals two standard deviations of the performance measured over the $2 * 3 = 6$ trained models.}
	\label{fig:ames_mutagenicity}
\end{figure*}

\begin{figure*}[h!]
	\centering
	\includegraphics[width=1.8\linewidth]{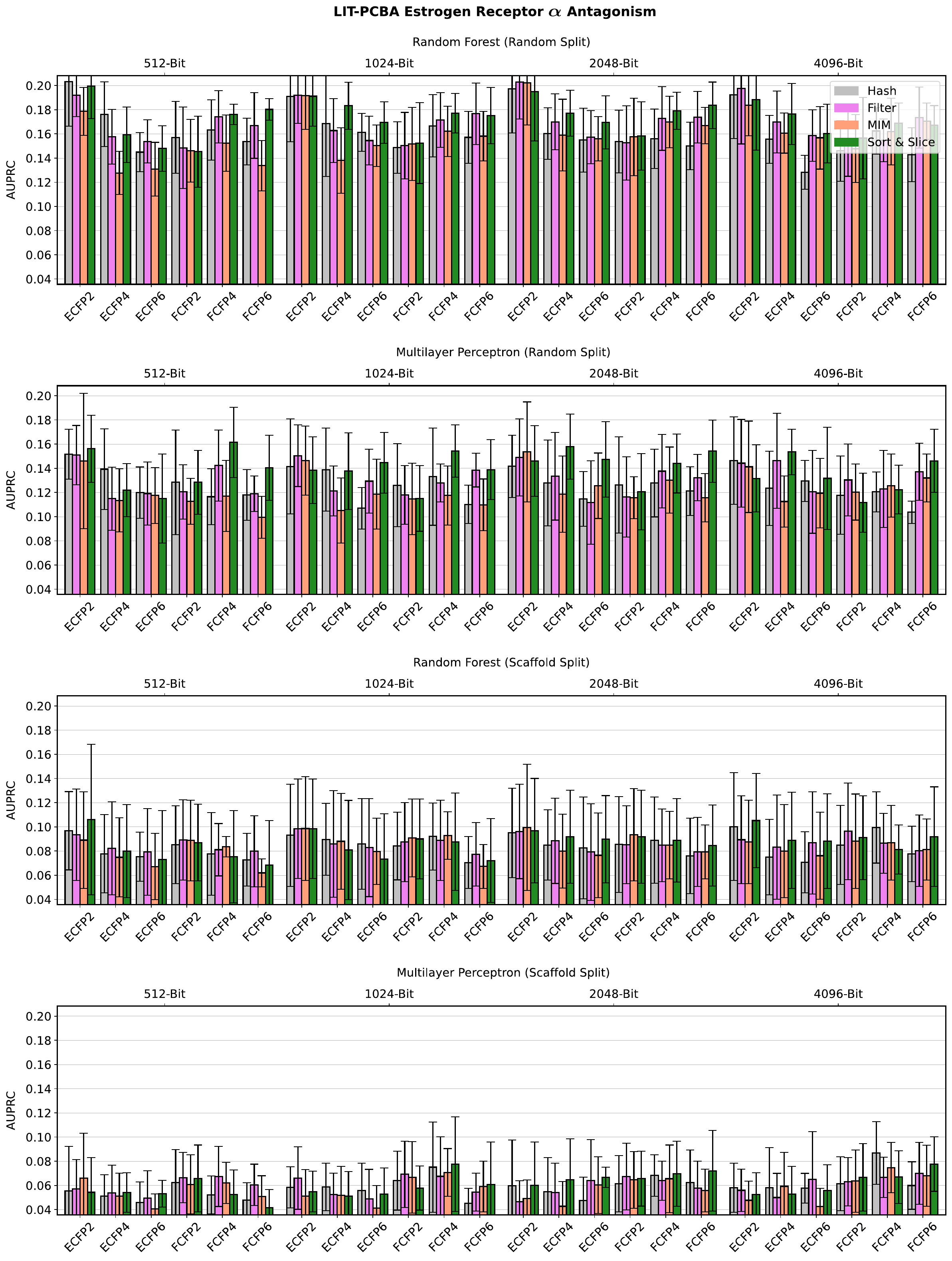}
	\caption{Predictive performance of the four investigated substructure-pooling methods (indicated by colours) for the highly imbalanced LIT-PCBA estrogen receptor $\alpha$ antagonism classification data set using varying data splitting techniques, machine-learning models and ECFP hyperparameters. Each bar shows the average area under the precision recall curve~(AUPRC) of the associated model across $2$-fold cross validation repeated with $3$ random seeds. The error bar length equals two standard deviations of the performance measured over the $2 * 3 = 6$ trained models.}
	\label{fig:lit_pcba_esr_ant}
\end{figure*}

In~\Cref{fig:aqsoldb_solubility,fig:postera_sars_cov_2_mpro,fig:ames_mutagenicity,fig:lit_pcba_esr_ant}, we present detailed computational results for the four remaining data sets specified in Table~\ref{tab: data_sets_substruc_pool}.



\begin{backmatter}
	
\section*{Acknowledgements} \justifying

\noindent We would like to thank Greg Landrum and Richard Cooper for their useful feedback.
	
\section*{Funding} \justifying
\noindent This research was supported by the University of Oxford's UK EPSRC Centre for Doctoral Training in Industrially Focused Mathematical Modelling (EP/L015803/1) and by the not-for-profit organisation and educational charity Lhasa Limited (\url{https://www.lhasalimited.org/}).

\section*{Abbreviations}

\begin{itemize}
	\item AUPRC = Area Under Precision Recall Curve
	\item AUROC = Area Under Receiver Operating Characteristic Curve
	\item ECFP = Extended-Connectivity Fingerprint
	\item FCFP = Functional-Connectivity Fingerprint
	\item GNN = Graph Neural Network
	\item MAE = Mean Absolute Error
	\item MIM = Mutual-Information Maximisation
	\item MLP = Multilayer Perceptron
	\item RF = Random Forest
	\item SMILES = Simplified Molecular-Input Line-Entry System
\end{itemize}

\section*{Availability of data and materials} \justifying

\noindent All used data sets and the \texttt{Python} code for our computational experiments are available in our public code repository
\url{https://github.com/MarkusFerdinandDablander/ECFP-substructure-pooling-Sort-and-Slice}.

\section*{Competing interests} \justifying
\noindent The authors declare that they have no competing interests.

\section*{Authors' contributions} \justifying

\noindent The computational study was designed, implemented, conducted and interpreted by the first author MD who also discovered and developed the described version of Sort \& Slice, the general mathematical framework, and the mathematical definition of substructure pooling. The research was supervised by GMM, TH and RL. The computer code was written by MD. The paper manuscript was written by MD. Feedback was provided by GMM, TH and RL during the writing process. All scientific figures were designed by MD, with input from GMM, TH and RL. When discussing how to name the investigated substructure-pooling technique, the name ``\textit{Sort \& Slice}" was suggested by GMM. All chemical data sets were gathered and cleaned by MD. All authors read and approved the final manuscript.


\bibliographystyle{spbasic_unsort} 
\bibliography{bmc_article}      



\end{backmatter}

\end{document}